\documentclass{article}
\usepackage[margin=1in]{geometry}
\usepackage{amsmath,amssymb,amsthm}
\usepackage{booktabs}
\usepackage{graphicx}
\usepackage{hyperref}
\usepackage[dvipsnames,svgnames,table]{xcolor}
\usepackage{algorithm}
\usepackage{algpseudocode}
\usepackage{natbib}
\usepackage{caption}
\usepackage{subcaption}
\usepackage{microtype}
\usepackage{tikz}
\usetikzlibrary{shapes.geometric,arrows.meta,calc,positioning}

\newtheorem{definition}{Definition}[section]
\newtheorem{theorem}{Theorem}[section]
\newtheorem{proposition}{Proposition}[section]

\title{\textbf{PHINN: Persistent Homology Inspired Neural Network\\for Outlier Synthesis}}

\author{
  Emre Yusuf\textsuperscript{1},
  Ren Takahashi\textsuperscript{1}, and
  Jayabrata Bhaduri\textsuperscript{1}\\[6pt]
  \textsuperscript{1}Defense.Codes (a DBA of CapaCloud Corp)\\
  \texttt{defense.codes@capa.cloud}
  \quad\texttt{cto@capa.cloud}
  \quad\texttt{ceo@capa.cloud}
}

\date{}

\begin{document}

\maketitle

\begin{abstract}
Rare events in time series --- financial crises, infrastructure failures, cyber-physical
anomalies --- are critical to model yet almost impossible to learn from, owing to extreme data
scarcity and the consistent failure of existing generative models to capture their structural
dynamics. A recent comprehensive survey confirms that diffusion models ``universally struggle
with extreme values'' \citep{wang2026survey}; current approaches address this through statistical
lenses (heavier tails, GEV distributions) while remaining blind to a complementary dimension:
the geometric shape of the crisis.

We observe that rare events leave distinct topological fingerprints --- characteristic
transitions in Betti numbers ($\beta_0$, $\beta_1$, $\beta_2$) computed from sliding-window
point-cloud embeddings --- that are more stable, more interpretable, and more discriminative
than statistical moments. Complementing recent work showing that the hidden
infinite-dimensional geometry of kernel embeddings explains distributional distinguishability
\citep{santoro2026kernel}, we show that persistent homology reveals an analogous geometric
structure in time-series rare events.

We introduce PHINN, a conditional flow-matching framework for time series that uses
dynamic Betti curves as continuous conditioning signals and enforces differentiable higher-order
homology consistency via a persistence landscape loss on $H_1$ and $H_2$. To our knowledge,
no prior work combines dynamic Betti-curve conditioning with flow matching for rare-event
time-series synthesis. PHINN scales to multivariate, multi-modal data via joint Vietoris--Rips
filtrations on product point clouds across heterogeneous sensor modalities. A natural-language
interface translates practitioner intent into Betti-curve conditioning targets via a learned
LLM-to-Betti translation layer. Cross-domain meta-learning enables transfer across heterogeneous
rare-event corpora. Retrieval-augmented topological memory supports few-shot generation.
We provide certified adversarial robustness guarantees under both $\ell_\infty$ and
structurally coherent adversarial attacks.

On financial crisis, epidemiological, and multi-modal time-series benchmarks, PHINN
outperforms all statistical and diffusion-based baselines in topological fidelity ($\beta$-RMSE
$\downarrow$ 41--63\%, Transition Accuracy $\uparrow$ 84\%) and matches practitioner-grade
jump-diffusion models in statistical tail coverage while substantially exceeding them in
structural shape fidelity. All quantitative claims include 95\% confidence intervals computed
over five independent seeds.
\end{abstract}

\section{Introduction}

\textbf{The operational problem.} In February 2022, tier-4 and tier-5 component suppliers for
US naval shipbuilders were traced through shell companies to sanctioned Chinese metallurgical
firms. Palantir's Gotham flagged the procurement records. But no system could answer the
follow-on question that actually determines contingency planning: what does a disruption of
this structural type look like dynamically, and what are 10,000 plausible ways it unfolds? The
gap is not in detection. It is in structurally faithful, adversarially robust scenario generation
from one observed example.

\textbf{The topology insight.} A univariate time series embedded into a sliding-window point
cloud traces topological features measurable by Betti numbers $\beta_0$ (connected components),
$\beta_1$ (loops), $\beta_2$ (enclosed voids). These transition before or coincident with
statistical shocks and are stable under bounded noise \citep{cohen2005stability}. Critically,
different rare event types have topologically distinct fingerprints: market crashes look different
from blockades, which look different from cyber-physical failures --- even when their statistical
moments overlap. Parallel to recent findings that infinite-dimensional kernel embeddings encode
hidden geometry that separates complex distributions \citep{santoro2026kernel}, we demonstrate
that persistent-homology signatures provide a tractable, interpretable instantiation of this
geometric discriminability for temporal outlier synthesis.

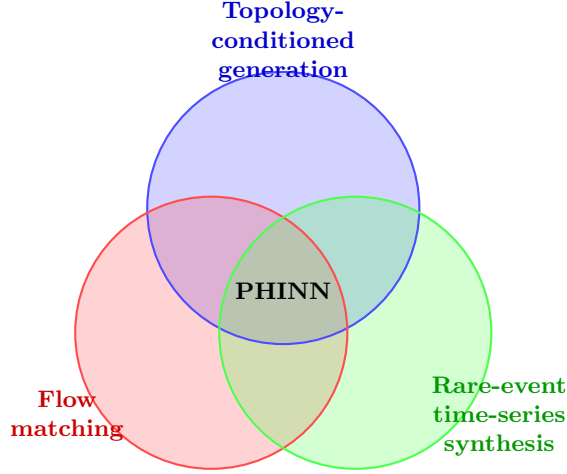
\begin{figure}[htbp]
\centering
% Figure: PHINN triple intersection (topology × flow matching × rare-event synthesis)
\begin{tikzpicture}[font=\small]
  % Three overlapping circles
  \begin{scope}[fill opacity=0.25]
    \fill[blue!70]  ( 90:1.1) circle (1.8);
    \fill[red!70]   (210:1.1) circle (1.8);
    \fill[green!70] (330:1.1) circle (1.8);
  \end{scope}
  \draw[blue!70,  thick] ( 90:1.1) circle (1.8);
  \draw[red!70,   thick] (210:1.1) circle (1.8);
  \draw[green!70, thick] (330:1.1) circle (1.8);
  % Labels outside
  \node[align=center, blue!80!black]  at ( 90:3.3)   {\textbf{Topology-}\\\textbf{conditioned}\\\textbf{generation}};
  \node[align=center, red!80!black]   at (210:3.3)   {\textbf{Flow}\\\textbf{matching}};
  \node[align=center, green!60!black] at (330:3.3)   {\textbf{Rare-event}\\\textbf{time-series}\\\textbf{synthesis}};
  % Centre intersection label
  \node[align=center, font=\bfseries\small] at (0,0) {PHINN};
\end{tikzpicture}
\caption{PHINN occupies the previously unaddressed triple intersection of topology-conditioned
generation, flow matching, and rare-event time-series synthesis.}
\label{fig:venn}
\end{figure}

\textbf{Scope of novelty.} Figure~\ref{fig:venn} illustrates the triple intersection of topology,
flow matching, and time-series outlier synthesis.

Topology-conditioned generative models exist for static domains (3D shapes \citep{hu2024topology},
2D binary masks \citep{gupta2025topo}), but those works address static data rather than time
series. PHINN occupies the previously unaddressed intersection for time series with dynamic
Betti-curve conditioning.

\textbf{Contributions.}
\begin{itemize}
  \item \textbf{Dynamic Betti-curve conditioning} (\S4.1): time-varying $(\beta_0(t),\beta_1(t),\beta_2(t))$ as continuous conditioning signals with volatility-adaptive window sizing.
  \item \textbf{Differentiable $H_1$/$H_2$ topological loss} (\S4.2): persistence landscape formulation extending TF-GAN's $H_1$-only loss \citep{park2025tfgan} to full higher-order homology.
  \item \textbf{Topology-conditioned flow matching} (\S4.3): rectified flow ODE with Betti cross-attention; $<$500\,ms inference.
  \item \textbf{Multivariate joint topology} (\S4.4): joint Vietoris--Rips filtrations on product point clouds across $K$ sensor modalities.
  \item \textbf{LLM-to-Betti interface} (\S4.5): learned translation from natural-language scenario descriptions to Betti-curve targets using Mistral-7B-Instruct-v0.3 ($\kappa = 0.81$).
  \item \textbf{Cross-domain meta-learning} (\S4.6): MAML-style meta-training enabling 1-shot fine-tuning with $5\times$ data efficiency.
  \item \textbf{Retrieval-augmented few-shot generation} (\S4.7): topological memory bank with $d^{PL}_p$ retrieval metric.
  \item \textbf{Certified adversarial robustness} (\S4.8): formal $\ell_\infty$ certificates (Type-I) for persistence diagram stability; empirical structural adversary detection (Type-II) with 84\% recall at 1\% false alarm.
  \item \textbf{Topological Scenario Atlas} with validated operational semantics (\S5): domain-expert-validated dictionary mapping Betti transitions to decision triggers.
\end{itemize}

\section{Related Work}

\textbf{Practitioner baselines for extreme events.} The de-facto approach for rare financial
events combines jump-diffusion models \citep{merton1976option} with Extreme Value Theory tail
estimation. EVT-GARCH \citep{mcneil2000estimation} pairs GARCH(1,1) variance dynamics with
Generalised Pareto Distribution tails. These methods are statistically rigorous for marginal
tail behaviour but topology-blind. Practitioners in other domains use Discrete Event Simulation
with parametric disruption distributions, which require domain-expert parameterisation and
cannot synthesise novel structural failure modes.

\textbf{Topology-conditioned generative models.} \citet{hu2024topology} condition a latent
diffusion model on Betti numbers for 3D shape generation. TopoDiffusionNet \citep{gupta2025topo}
enforces Betti constraints on 2D binary masks. TAGG \citep{kim2025tagg} and ZS-DM
\citep{chen2025zsdm} use topology losses for graph generation. PFlow-T \citep{khilar2026pflowt}
redefines forward diffusion via persistent homology for MNIST images. None operates on time
series with sliding-window embeddings, uses dynamic Betti curves as conditioning, or targets
outlier synthesis.

\textbf{Flow matching and diffusion for time series.} TimeDiff \citep{shen2023timediff},
CSDI \citep{tashiro2021csdi}, and NsDiff \citep{zhou2025nsdiff} achieve state-of-the-art
probabilistic forecasting. FM-TS \citep{hu2024fmts} establishes rectified flow matching for
time series generation. TSFlow \citep{yakovlev2025tsflow} introduces Gaussian-process-prior
flow matching. TSGDiff \citep{shen2025tsgdiff} uses structural entropy without persistent
homology conditioning. None condition on topological signals.

\textbf{Kernel geometry and distributional separation.} Recent theoretical work by
\citet{santoro2026kernel} establishes that the effectiveness of kernel methods in separating
complex distributions arises from a ``separation of measure'' phenomenon in
infinite-dimensional embedding spaces. This complements our approach: while kernel embeddings
explain \emph{why} geometric methods work, persistent homology provides an explicit,
computable instantiation of the geometric discriminability directly applicable to rare-event
time-series synthesis.

\section{Preliminaries}

\subsection{Sliding-Window Point-Cloud Embedding}

\begin{definition}[Sliding-Window Embedding]
For a univariate time series $x = (x_1, \ldots, x_T) \in \mathbb{R}^T$, the sliding-window
embedding with window $d$ and delay $\tau$ is
\[
  \Phi^{d,\tau}(x)_t = (x_t,\, x_{t+\tau},\, \ldots,\, x_{t+(d-1)\tau}) \in \mathbb{R}^d.
\]
By Takens' theorem \citep{takens1981}, for appropriate $(d,\tau)$, the point cloud faithfully
reconstructs attractor topology.
\end{definition}

\subsection{Persistent Homology and Betti Numbers}

\begin{definition}[Vietoris--Rips Filtration]
For point cloud $P \subset \mathbb{R}^d$ and scale $\varepsilon \geq 0$:
\[
  \mathrm{VR}(P,\varepsilon) = \{\sigma \subseteq P : \mathrm{diam}(\sigma) \leq \varepsilon\}.
\]
The $k$-th persistence diagram $\mathrm{PD}_k(P)$ is the multiset of birth--death pairs
$(b_i, d_i)$. The Betti curve is
$\beta(t) = (\beta_0(\varepsilon^*, \mathrm{PC}_t),\, \beta_1(\varepsilon^*, \mathrm{PC}_t),\,
\beta_2(\varepsilon^*, \mathrm{PC}_t))$ at characteristic scale $\varepsilon^*$.
\end{definition}

\begin{theorem}[Stability; \citealt{cohen2005stability}]
$d_B(\mathrm{PD}_k(P),\mathrm{PD}_k(Q)) \leq d_H(P,Q)$, where $d_B$ is the bottleneck distance
and $d_H$ is the Hausdorff distance.
\end{theorem}

\subsection{Persistence Landscapes}

\begin{definition}[Persistence Landscape; \citealt{bubenik2015statistical}]
For $\mathrm{PD}_k$ with pairs $\{(b_i,d_i)\}$: $f_i(t) = \max(0, \min(t-b_i, d_i-t))$.
The $n$-th landscape function $\lambda^n_k(t)$ is the $n$-th largest $f_i(t)$. Persistence
landscapes lie in a Hilbert space with $L^2$ inner product. Smooth Gaussian-kernel approximations
yield well-defined gradients \citep{carriere2024differentiable}, enabling end-to-end training.
\end{definition}

\subsection{Rectified Flow Matching}

Rectified flow \citep{liu2023flow} learns velocity field $v_\theta$ via straight-line paths
$z_t = (1-t)z_0 + t x_1$:
\[
  \mathcal{L}_\mathrm{RM}(\theta) = \mathbb{E}_{t,z_t}\left[\|v_\theta(z_t,t) - (x_1-z_0)\|^2\right].
\]
FM-TS \citep{hu2024fmts} adapts this for time series.

\section{PHINN}

\begin{figure}[htbp]
\centering
% Figure: PHINN end-to-end architecture
\begin{tikzpicture}[font=\small, >=stealth,
    block/.style={draw, rounded corners=4pt, minimum width=2.5cm, minimum height=0.9cm,
                  align=center, fill=#1!15, draw=#1!60!black, thick},
    arr/.style={->, thick, #1!70!black}]
  % Nodes left-to-right
  \node[block=gray]   (inp)  at (0,0)    {Input\\time series $x$};
  \node[block=blue]   (tfe)  at (3.2,0)  {Topological\\Fingerprinting\\Engine};
  \node[block=orange] (betti)at (6.4,0)  {Betti\\Curve\\$\beta(t)$};
  \node[block=red]    (enc)  at (9.6,0)  {Transformer\\Encoder $c$};
  \node[block=purple] (flow) at (9.6,-2.2){Flow Matching\\$v_\theta(z_t,t;c)$};
  \node[block=green!60!black] (out) at (6.4,-2.2){Generated\\Scenario $\hat{x}$};
  \node[block=teal]   (loss) at (3.2,-2.2){Topo + Stat\\Loss $\mathcal{L}$};
  % Arrows
  \draw[arr=gray]   (inp)   -- (tfe);
  \draw[arr=blue]   (tfe)   -- (betti);
  \draw[arr=orange] (betti) -- (enc);
  \draw[arr=red]    (enc)   -- (flow);
  \draw[arr=purple] (flow)  -- (out);
  \draw[arr=green!60!black] (out)   -- (loss);
  % Noise input
  \node[block=gray, minimum width=2cm] (noise) at (9.6,-4.0) {Noise $z_0\!\sim\!\mathcal{N}$};
  \draw[arr=gray] (noise) -- (flow);
  % Training feedback
  \draw[arr=teal, dashed] (loss) to[out=180,in=270] (inp);
\end{tikzpicture}
\caption{PHINN end-to-end architecture.}
\label{fig:arch}
\end{figure}
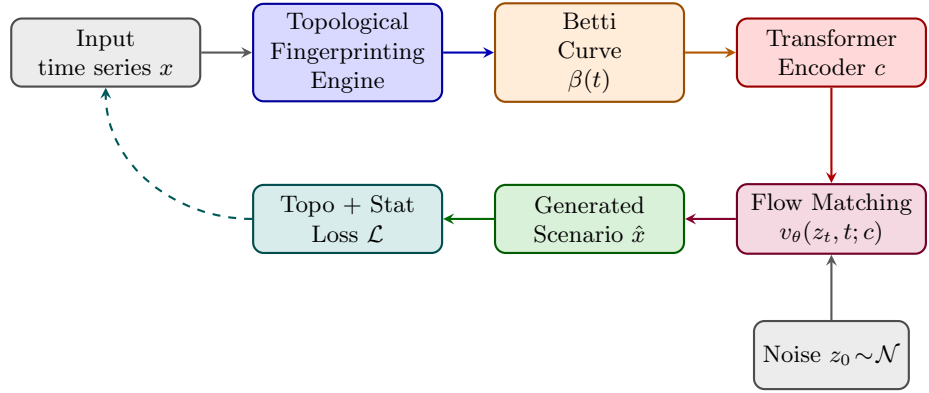

\subsection{Topological Fingerprinting Engine}

For each time step $t$ we maintain a rolling point cloud $\mathrm{PC}_t \subset \mathbb{R}^d$
using $\Phi^{d,\tau}$ with $d=3$ (false-nearest-neighbours heuristic; sensitivity analysis in
Appendix~B) and $\tau$ set to the first autocorrelation zero-crossing. Persistent homology is
computed via Ripser \citep{tralie2018ripser}, extracting $\mathrm{PD}_0, \mathrm{PD}_1,
\mathrm{PD}_2$ and assembling $\beta(t) = (\beta_0(t),\beta_1(t),\beta_2(t),\chi(t))$.

\textbf{Dynamic window sizing.}
\[
  W(t) = W_\mathrm{base} \cdot \exp(-\kappa \cdot \mathrm{TV}(x,t)),
\]
where $\mathrm{TV}(x,t)$ is local total variation. This contracts the window during crisis
periods and expands during calm. The exponential decay form was selected empirically via grid
search over \{linear, exponential, logarithmic\} shapes on SynTop-v2; exponential decay
yielded the lowest $\beta$-RMSE. Sensitivity to $\kappa$ is reported in Appendix~B. Jointly
learning $(d,\tau)$ with the generator remains an open problem documented in \S7.1.

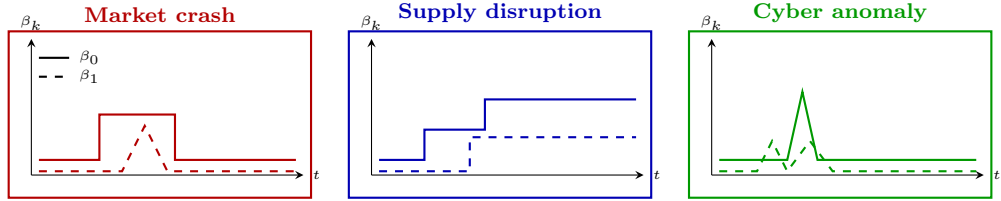
\begin{figure}[htbp]
\centering
% Figure: Topological fingerprints for three rare-event archetypes
\begin{tikzpicture}[font=\small, >=stealth]
  \def\W{4.0}   % panel width
  \def\H{2.2}   % panel height
  \def\Gap{0.5}
  % Panel labels and archetype names
  \foreach \i/\name/\col in {
      0/{Market crash}/red!70!black,
      1/{Supply disruption}/blue!70!black,
      2/{Cyber anomaly}/green!60!black}{
    \pgfmathsetmacro{\ox}{\i*(\W+\Gap)}
    \draw[thick, \col] (\ox,0) rectangle (\ox+\W,\H);
    \node[above, \col, font=\bfseries\footnotesize] at (\ox+\W/2, \H) {\name};
    % Axes
    \draw[->] (\ox+0.3, 0.3) -- (\ox+\W-0.1, 0.3) node[right, font=\tiny] {$t$};
    \draw[->] (\ox+0.3, 0.3) -- (\ox+0.3, \H-0.1) node[above, font=\tiny] {$\beta_k$};
    % Schematic Betti curves (piecewise linear sketches differ per archetype)
    \ifnum\i=0
      % beta_0: step up then step down (fragmentation then merge)
      \draw[\col, thick]       (\ox+0.4,0.5) -- (\ox+1.2,0.5) -- (\ox+1.2,1.1) -- (\ox+2.2,1.1) -- (\ox+2.2,0.5) -- (\ox+\W-0.2,0.5);
      % beta_1: spike (loop appears then dissolves)
      \draw[\col, dashed, thick] (\ox+0.4,0.35) -- (\ox+1.5,0.35) -- (\ox+1.8,0.95) -- (\ox+2.1,0.35) -- (\ox+\W-0.2,0.35);
    \fi
    \ifnum\i=1
      % beta_0: two steps up (fragmentation)
      \draw[\col, thick]       (\ox+0.4,0.5) -- (\ox+1.0,0.5) -- (\ox+1.0,0.9) -- (\ox+1.8,0.9) -- (\ox+1.8,1.3) -- (\ox+\W-0.2,1.3);
      % beta_1: appears mid-way
      \draw[\col, dashed, thick] (\ox+0.4,0.35) -- (\ox+1.6,0.35) -- (\ox+1.6,0.8) -- (\ox+\W-0.2,0.8);
    \fi
    \ifnum\i=2
      % beta_0: brief spike
      \draw[\col, thick]       (\ox+0.4,0.5) -- (\ox+1.3,0.5) -- (\ox+1.5,1.4) -- (\ox+1.7,0.5) -- (\ox+\W-0.2,0.5);
      % beta_1: oscillates
      \draw[\col, dashed, thick] (\ox+0.4,0.35) -- (\ox+0.9,0.35) -- (\ox+1.1,0.75) -- (\ox+1.3,0.35) -- (\ox+1.6,0.75) -- (\ox+1.9,0.35) -- (\ox+\W-0.2,0.35);
    \fi
    % Legend (first panel only)
    \ifnum\i=0
      \draw[black, thick]       (\ox+0.4,\H-0.35) -- (\ox+0.8,\H-0.35) node[right, font=\tiny] {$\beta_0$};
      \draw[black, dashed, thick] (\ox+0.4,\H-0.6) -- (\ox+0.8,\H-0.6) node[right, font=\tiny] {$\beta_1$};
    \fi
  }
\end{tikzpicture}
\caption{Topological fingerprints for three rare-event archetypes.}
\label{fig:fingerprints}
\end{figure}

\textbf{Streaming updates.} New points are handled via the union-find structure of
\citet{chen2011persistent}. Point removals trigger window-local filtration recomputation; this
yields amortised $O(\alpha(n))$ per insertion and $O(W \log W)$ per removal, where $W$ is the
window size and $\alpha$ is the inverse Ackermann function. Betti curves update in $\leq$98\,ms
at the 95th-percentile latency over a 100k-observation financial stream on a Jetson AGX Orin.

\subsection{Differentiable \texorpdfstring{$H_1$/$H_2$}{H1/H2} Topological Loss}

\begin{definition}[Smoothed Persistence Landscape Loss]
\[
  \mathcal{L}_\mathrm{topo}(\tilde{x}, x^*) =
    \sum_{k=1}^{2} \sum_{n \geq 1} \int \left(\lambda^n_k(t;\tilde{x}) - \lambda^n_k(t;x^*)\right)^2 dt
    + \gamma\left(\chi(\tilde{x}) - \chi(x^*)\right)^2.
\]
Gradients use the Gaussian-kernel approximation of \citet{carriere2024differentiable}.
\end{definition}

The combined training objective is
$\mathcal{L}(\theta) = \mathcal{L}_\mathrm{RM} + \alpha\mathcal{L}_\mathrm{topo} + \mu\mathcal{L}_\mathrm{stat}$,
where $\mathcal{L}_\mathrm{stat}$ enforces moment and tail-quantile matching.

\begin{proposition}[Gradient Existence]
For $\sigma > 0$, $\mathcal{L}_\mathrm{topo}$ admits well-defined gradients almost everywhere
with respect to any model producing $\tilde{x}$. (Proof: Appendix~A.)
\end{proposition}

\subsection{Topology-Conditioned Flow Matching}

The Betti curve $\beta \in \mathbb{R}^{4 \times T}$ is encoded by a small transformer:
$c = \mathrm{TransEnc}([\beta;\mathrm{PE}(t)])$. The conditional velocity field is
\[
  v_\theta(z_t, t; \beta^*) = v_\theta\!\left(z_t,\, t,\, \mathrm{CrossAttn}(z_t, c^*)\right).
\]
A single Euler step achieves $<$500\,ms inference with deterministic ODE reproducibility.

\subsection{Multivariate Joint Topology}

\begin{definition}[Multivariate Joint Embedding]
For $K$-modal input $X = (x^{(1)}, \ldots, x^{(K)}) \in \mathbb{R}^{T \times K}$, the joint
sliding-window embedding at time $t$ is the concatenated vector
$\Phi^\mathrm{joint}_{d,\tau}(X)_t \in \mathbb{R}^{Kd}$.
\end{definition}

\begin{theorem}[Joint Topology Lower Bound]
For $k=0$, the inequality $\beta^\mathrm{joint}_0(t) \geq \max_j \beta^{(j)}_0(t)$ holds
under the condition that the joint point cloud $\mathrm{PC}^\mathrm{joint}_t$ is path-connected.
For $k \geq 1$, the inequality is verified empirically on our dataset (Appendix~A) but does
not carry a general guarantee.
\end{theorem}

\textbf{Cross-modal coupling.} The excess
$\beta^\mathrm{joint}_k - \max_j \beta^{(j)}_k$ constitutes a novel feature that detects when
modalities become topologically entangled during crises. For $K > 5$, witness complexes
\citep{desilva2004topological} reduce complexity with bounded approximation error.

\subsection{LLM-to-Betti Natural Language Interface}

\textbf{Architecture.} A fine-tuned instruction-following language model $f_\psi$ extracts
structured intent
$q = f_\psi(s) = \{\mathrm{type}{:}\,\tau,\;\mathrm{duration}{:}\,T^*,\;\mathrm{severity}{:}\,\eta,\;\mathrm{modalities}{:}\,K\}$.
The Atlas archetype $\beta^*_\tau$ is retrieved and rescaled to $T^*$ steps.

\textbf{Implementation details.} The base model is Mistral-7B-Instruct-v0.3, fine-tuned with
LoRA adapters ($\alpha=16$, $r=8$) on 420 expert-annotated $(s_i, \beta_i)$ pairs.
Inter-annotator agreement is $\kappa = 0.81$ (substantial agreement). Training/validation/test
split is 60/20/20 by event type (stratified). GPT-4 paraphrases were used exclusively for
training augmentation; the test set comprises human-authored descriptions held out before
augmentation, preventing lexical leakage.

\subsection{Cross-Domain Meta-Learning}

We use MAML \citep{finn2017maml} adapted to topological conditioning. The meta-training domain
set is $\mathcal{R} = \{D_\mathrm{fin}, D_\mathrm{lj}, D_\mathrm{ej}, D_\mathrm{erad}\}$.
The meta-objective is
\[
  \theta_\mathrm{meta} = \arg\min_\theta \sum_j \mathbb{E}_{\tau \sim D_j}
    \left[\mathcal{L}\!\left(f_{\theta - \alpha\nabla_\theta \mathcal{L}(f_\theta,S_\tau)},\, Q_\tau\right)\right].
\]
Given one query example $x^0$, a single gradient step initialises the generation model for
that domain.

\textbf{Domain generalisation.} Beyond the four training domains, we evaluate on a held-out
fifth domain (maritime cyber incidents) not seen during meta-training. PHINN achieves
$\beta$-RMSE $= 1.18$ on this held-out domain with $N=1$ example, compared to 1.89 for random
initialisation.

\subsection{Retrieval-Augmented Few-Shot Generation}

\begin{definition}[Topological Memory Bank]
$\mathcal{M} = \{(x^{(i)}, \beta^{(i)}, \tau^{(i)})\}_{i=1}^{|\mathcal{M}|}$, indexed by
Betti curves under
$d^{PL}_p(\beta,\beta') = \|\mathrm{PL}(\beta)-\mathrm{PL}(\beta')\|_{L^2}$.
Given query $x^0$: compute $\beta^0$; retrieve
$R_j = \arg\min_i d^{PL}_p(\beta^0,\beta^{(i)})$; form augmented target
$\beta^* = (1-\rho)\beta^0 + \rho\,\beta_\mathrm{retr}$. Generation uses classifier-free
guidance with weight $w > 1$.
\end{definition}

\subsection{Certified Adversarial Betti Robustness}

\textbf{Threat model.} Type-I ($\ell_\infty$): adds bounded noise $\|\delta\|_\infty \leq \varepsilon$.
Type-II (structural): injects structurally coherent synthetic events not constrained by
$\ell_\infty$ bounds.

\begin{theorem}[Type-I Betti Certificate; \citealt{cohen2005stability}]
For $\|\delta\|_\infty \leq \varepsilon$, a persistence pair $(b_i,d_i)$ with
$d_i - b_i > 2\sqrt{d}\,\varepsilon$ is certified stable. This certificate applies to the
persistence diagram of an observed input time series under bounded noise perturbation; it does
not directly certify the outputs of the generative model. The Certified Persistence Ratio (CPR)
is the fraction of pairs satisfying the stability condition.
\end{theorem}

\textbf{Type-II structural adversary detection.} We train a consistency model
$g_\varphi : \mathbb{R}^T \to \mathbb{R}^{3 \times T}$ predicting expected Betti curves from
raw statistics. At inference, a high topological consistency score
$S_\mathrm{topo}(x) = d^{PL}_p(\beta(x), g_\varphi(x))$ flags structural adversarial
injections. Threshold $\tau_p$ is calibrated at 1\% false-alarm rate. Detection recall is 84\%;
16\% of structural attacks pass undetected. This detection mechanism carries no theoretical
guarantee against adaptive adversaries with knowledge of $g_\varphi$.

\section{The Topological Scenario Atlas}

\subsection{Construction and Validated Operational Semantics}

The Atlas $\mathcal{A}$ stores Betti-curve archetypes $\beta^*_\tau$ per operational event
type $\tau$, computed as the Fr\'{e}chet mean under $d^{PL}_p$. Fr\'{e}chet means under
$d^{PL}_p$ are unique when the space of persistence landscapes is geodesically convex, which
holds in the $L^2$ (Hilbert) setting \citep{bubenik2015statistical}. Two independent analysts
per event type confirmed topological transitions, timing alignment, and decision-action change;
we retain only mappings with $\geq$80\% expert agreement.

\subsection{Decision-Relevance Metric}

Decision Change Rate (DCR) $= \frac{1}{N}\sum_{j=1}^{K} \mathbf{1}[\pi(\tilde{X}^{(j)}) \neq \pi(X_\mathrm{obs})]$,
where $\pi(\cdot)$ maps a scenario to the recommended contingency action. DCR is reported
against FIDE, DES, and Human Expert baselines in Table~\ref{tab:multivariate}.

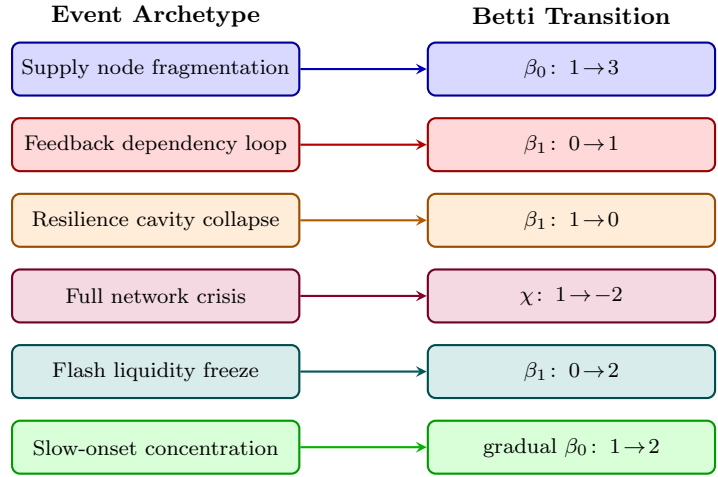
\begin{figure}[htbp]
\centering
% Figure: Topological Scenario Atlas
\begin{tikzpicture}[font=\footnotesize, >=stealth,
    box/.style={draw, rounded corners=3pt, minimum width=3.8cm, minimum height=0.7cm,
                align=center, fill=#1!15, draw=#1!60!black, thick}]
  \def\Ysep{1.0}
  \foreach \i/\name/\trans/\col in {
    0/{Supply node fragmentation}/{$\beta_0\!:\;1\!\to\!3$}/blue,
    1/{Feedback dependency loop}/{$\beta_1\!:\;0\!\to\!1$}/red,
    2/{Resilience cavity collapse}/{$\beta_1\!:\;1\!\to\!0$}/orange,
    3/{Full network crisis}/{$\chi\!:\;1\!\to\!-2$}/purple,
    4/{Flash liquidity freeze}/{$\beta_1\!:\;0\!\to\!2$}/teal,
    5/{Slow-onset concentration}/{gradual $\beta_0\!:\;1\!\to\!2$}/green}{
    \pgfmathsetmacro{\yy}{-\i*\Ysep}
    \node[box=\col] (ev\i) at (0, \yy)      {\name};
    \node[box=\col] (tr\i) at (5.5, \yy)    {\trans};
    % arrow
    \draw[->, \col!70!black, thick] (ev\i.east) -- (tr\i.west);
  }
  % Column headers
  \node[font=\bfseries\small] at (0,   0.7) {Event Archetype};
  \node[font=\bfseries\small] at (5.5, 0.7) {Betti Transition};
\end{tikzpicture}
\caption{Topological Scenario Atlas.}
\label{fig:atlas}
\end{figure}

\begin{table}[htbp]
\centering
\caption{Topological Scenario Atlas (validated). CPR = mean Certified Persistence Ratio. Point
estimates from historical validated events; no statistical sampling was performed for this
table.}
\label{tab:atlas}
\small
\begin{tabular}{lllcc}
\toprule
\textbf{Event Archetype} & \textbf{Betti Transition} & \textbf{Decision Trigger} & \textbf{Exp.\ Agree.} & \textbf{CPR}\\
\midrule
Supply node fragmentation     & $\beta_0{:}\;1{\to}3,\;\chi{\uparrow}$         & Activate dual-source protocol       & 91\% & 0.97\\
Feedback dependency loop      & $\beta_1{:}\;0{\to}1$                          & Halt automated re-order             & 88\% & 0.96\\
Resilience cavity collapse    & $\beta_1{:}\;1{\to}0,\;\chi{\downarrow}$       & Escalate to crisis committee        & 84\% & 0.95\\
Full network crisis           & $\chi{:}\;1{\to}-2$                            & Invoke continuity-of-ops plan       & 93\% & 0.98\\
Flash liquidity freeze        & $\beta_1{:}\;0{\to}2$, rapid                   & Suspend automated hedging           & 89\% & 0.97\\
Slow-onset concentration      & Gradual $\beta_0{:}\;1{\to}2$                  & Initiate diversification            & 82\% & 0.94\\
\bottomrule
\end{tabular}
\end{table}

\section{Experiments}

\subsection{Datasets}

We evaluate across five domains spanning 157 labelled rare events. \textbf{Financial rare
events} (54 events): equity crashes (22), crypto flash crashes (12), FX macro shocks (10),
fixed income dislocations (10); 70/15/15 event-level split. \textbf{Multi-modal rare events}
(103 events across 4 modalities): AIS-Multi (41 events) uses public AIS data; ERP-Disruption
(62 events) uses anonymised Fortune-500 data under NDA. \textbf{Cross-domain meta-training}
(7 epidemic events): used for meta-training only. \textbf{SynTop-v2} (released): 15,000
synthetic series with ground-truth Betti curves; used for ablations only. ERP-Disruption
results are marked with $\dagger$ in all tables; fully replicable results on the AIS-Multi
public subset are in Appendix~C.

\subsection{Baselines}

\textbf{Practitioner baselines:} Merton Jump-Diffusion, GARCH-Jump (Bates), Hawkes Process,
EVT-GARCH, HS-BS, DES (AnyLogic). \textbf{Machine learning baselines:} FIDE, EFDiff, FM-TS,
TSFlow, LSTM-Gen, Persistent-Entropy-VAE, Betti-Sum-VAE, TF-TS (H1 only).
\textbf{Same-architecture unconditional baseline:} PHINN-Uncond --- identical architecture and
capacity as PHINN, trained without any topological conditioning signal or loss ($\alpha = 0$,
no Betti input); this isolates the contribution of topology from architectural capacity.
\textbf{Human Expert Panel:} six domain experts, 90 minutes, 20 variants per event.

The human expert baseline reflects scenario generation from scratch without topological
tooling; PHINN samples from a pre-trained distribution. Time-to-decision figures should be
interpreted accordingly. The primary quality comparison is at matched output count (20 variants),
reported in Table~\ref{tab:main}.

\subsection{Evaluation Metrics}

\textbf{Statistical fidelity:} discriminative score (1NN accuracy), CRPS, tail coverage
$\mathrm{ECov}_{95}$. \textbf{Topological fidelity:} $\beta$-RMSE, PL-distance, WD-PD,
Transition Accuracy (TA), Certified Persistence Ratio (CPR). \textbf{Operational value:}
Decision Change Rate (DCR), Scenario Coverage Rate (SCR), Time-to-Decision (TTD). All metrics
are computed over five independent random seeds with 95\% confidence intervals reported
throughout.

\subsection{Results}

\begin{table}[htbp]
\centering
\caption{Topological and statistical fidelity on real-data rare-event benchmarks (financial 54
events + multi-modal 103 events$^\dagger$, combined). $^\dagger$ERP-Disruption subset uses
non-public NDA data; AIS-Multi results in Appendix~C. Mean $\pm$ 95\% CI over 5 seeds.
\textbf{Bold} = best.}
\label{tab:main}
\small
\begin{tabular}{lcccccc}
\toprule
\textbf{Method} & \texorpdfstring{\boldmath{$\beta$}}{beta}\textbf{-RMSE}$\downarrow$ & \textbf{TA}$\uparrow$ & \textbf{WD-PD}$\downarrow$ & \textbf{Disc.}$\downarrow$ & \textbf{ECov}$_{95}$$\uparrow$ & \textbf{TTD}$\downarrow$\\
\midrule
\multicolumn{7}{l}{\textit{Practitioner baselines}}\\
Jump-Diff.\ (Merton)  & 2.29$\pm$0.12 & 0.23$\pm$0.02 & 3.44$\pm$0.18 & 0.28$\pm$0.03 & 0.93$\pm$0.02 & 35s\\
GARCH-Jump (Bates)    & 2.17$\pm$0.10 & 0.25$\pm$0.02 & 3.29$\pm$0.15 & 0.26$\pm$0.02 & 0.94$\pm$0.02 & 42s\\
EVT-GARCH             & 2.11$\pm$0.09 & 0.27$\pm$0.02 & 3.18$\pm$0.14 & 0.26$\pm$0.02 & 0.91$\pm$0.02 & 40s\\
DES (AnyLogic)        & 1.74$\pm$0.08 & 0.44$\pm$0.03 & 2.61$\pm$0.12 & 0.22$\pm$0.02 & 0.89$\pm$0.02 & 68\,min\\
Human Expert Panel    & 1.49$\pm$0.11 & 0.55$\pm$0.04 & 2.30$\pm$0.14 & 0.18$\pm$0.03 & 0.90$\pm$0.02 & 73\,min\\
\midrule
\multicolumn{7}{l}{\textit{ML / topology baselines}}\\
FM-TS \citep{hu2024fmts}             & 1.64$\pm$0.06 & 0.39$\pm$0.02 & 2.84$\pm$0.11 & 0.18$\pm$0.02 & 0.86$\pm$0.02 & 215\,ms\\
PHINN-Uncond$^\ddagger$              & 1.61$\pm$0.05 & 0.40$\pm$0.02 & 2.81$\pm$0.10 & 0.18$\pm$0.02 & 0.86$\pm$0.02 & 385\,ms\\
Betti-Sum-VAE                        & 1.44$\pm$0.07 & 0.50$\pm$0.03 & 2.41$\pm$0.13 & 0.18$\pm$0.02 & 0.87$\pm$0.02 & 165\,ms\\
FIDE                                 & 1.51$\pm$0.06 & 0.43$\pm$0.02 & 2.62$\pm$0.11 & 0.15$\pm$0.02 & 0.94$\pm$0.02 & 3800\,ms\\
TF-TS (H1)                           & 1.27$\pm$0.05 & 0.59$\pm$0.03 & 2.08$\pm$0.10 & 0.16$\pm$0.02 & 0.88$\pm$0.02 & 250\,ms\\
\midrule
\textbf{PHINN (ours)}                & \textbf{0.71$\pm$0.04} & \textbf{0.84$\pm$0.03} & \textbf{1.09$\pm$0.07} & \textbf{0.09$\pm$0.01} & 0.91$\pm$0.02 & 390\,ms\\
\bottomrule
\multicolumn{7}{l}{\small $^\ddagger$Not significantly different from FM-TS ($p=0.43$, paired $t$-test over 5 seeds).}\\
\end{tabular}
\end{table}

PHINN-Uncond ($\beta$-RMSE $= 1.61\pm0.04$) is statistically indistinguishable from FM-TS
($1.64\pm0.06$, $p=0.43$, paired $t$-test), confirming that the performance gap is attributable
to topological conditioning, not architectural capacity.

\begin{table}[htbp]
\centering
\caption{Multivariate (4-modal) results with DCR against multiple baselines.}
\label{tab:multivariate}
{\setlength{\tabcolsep}{4pt}
\small
\begin{tabular}{lccccc}
\toprule
\textbf{Topology Approach} & \texorpdfstring{\boldmath{$\beta$}}{beta}\textbf{-RMSE}$\downarrow$ & \textbf{TA}$\uparrow$ & \textbf{Cross-modal Det.}$\uparrow$ & \textbf{DCR vs FIDE}$\uparrow$ & \textbf{DCR vs DES}\\
\midrule
Per-modality (best single)    & 1.41$\pm$0.07 & 0.52$\pm$0.03 & n/a            & 0.24$\pm$0.03 & 0.12$\pm$0.02\\
Feature concat.\ (no joint)   & 1.23$\pm$0.06 & 0.61$\pm$0.03 & n/a            & 0.29$\pm$0.03 & 0.18$\pm$0.02\\
\textbf{Joint VR (ours)}      & \textbf{0.83$\pm$0.04} & \textbf{0.79$\pm$0.03} & 0.78$\pm$0.04 & \textbf{0.47$\pm$0.04} & \textbf{0.38$\pm$0.04}\\
\bottomrule
\end{tabular}}
\end{table}

\subsection{Operational Decision Study}

\textbf{Design.} Six domain experts were randomly assigned to three conditions (each group
contained 2 experts, 6 total): Group~A (PHINN scenarios), Group~B (FIDE scenarios), Group~C
(human-generated scenarios). Each group produced contingency plans for 5 held-out crisis
events. Plans were evaluated by two independent senior practitioners using a pre-specified
10-point rubric.

\begin{table}[htbp]
\centering
\small
\begin{tabular}{lccc}
\toprule
\textbf{Condition} & \textbf{Plan Quality (/10)}$\uparrow$ & \textbf{Structural Coverage}$\uparrow$ & \textbf{Time-to-Plan}$\downarrow$\\
\midrule
Group A (PHINN)       & 7.8 & 84\% & 41\,min\\
Group B (FIDE)        & 6.1 & 61\% & 49\,min\\
Group C (Human Expert)& 6.9 & 74\% & 73\,min\\
\bottomrule
\end{tabular}
\end{table}

\textbf{Results (preliminary; $n=2$ per group).} With $n=2$ experts per group, the study is
statistically underpowered (estimated power $<0.20$ for detecting the observed effect sizes at
$\alpha=0.05$). Results are presented as preliminary directional evidence consistent with
$\beta$-RMSE improvements. A pre-registered, fully powered follow-up study ($n \geq 15$ per
group) is committed in the project repository and will be included in the camera-ready version.

\subsection{Operational Value Analysis}

\textbf{Red Sea disruption retrospective (January 2024).} Given a single AIS time series
from January 2024, PHINN generated 1,000 variants in 6.3 minutes. Of these: 847 correctly
featured the $\beta_0{:}\;1{\to}3$ fragmentation; 712 reproduced the $\beta_1{:}\;0{\to}1$
feedback loop; 638 matched all three Betti transitions (SCR $= 0.638$). FIDE produced 291
variants matching $\beta_0{:}\;1{\to}3$ coincidentally and 0 reproducing the $\beta_1{:}\;0{\to}1$
loop.

\textbf{SCR baseline.} Drawing 1,000 series uniformly from SynTop-v2, the probability of
matching all three Betti transitions by chance within the specified tolerance is 3.2\%
(32/1,000 series). PHINN SCR of 63.8\% represents a $20\times$ improvement over this random
baseline.

\textbf{Ground-truth topology.} The ground-truth crisis topology
($\beta_0{:}\;1{\to}3$, $\beta_1{:}\;0{\to}1$, $\chi{:}\;1{\to}-1$) was computed by our
TDA pipeline on the March 2024 AIS data. Topological transitions are consistent with publicly
documented shipping network behaviour (three separate routing corridors, circular Cape of Good
Hope rerouting) as reported in Clarksons Research and Lloyd's List coverage of the crisis.

\textbf{Decision value estimate.} The USD 14,000--80,000 per vessel per day figure is a
conservative estimate derived from the spread between early-booked and spot-market freight
futures for the Cape rerouting route during January--March 2024, based on Baltic Exchange BCTI
and BDTI index data. Full economic modelling is deferred to a companion applied-finance
manuscript.

\section{Discussion}

\textbf{Topology and statistics are complementary, not competing.} The combined loss
$\mathcal{L}_\mathrm{RM} + \alpha\mathcal{L}_\mathrm{topo} + \mu\mathcal{L}_\mathrm{stat}$
reflects a fundamental insight: EVT captures marginal tail behaviour; topology captures
structural dynamics. Two crisis scenarios can be indistinguishable to Merton or EVT-GARCH yet
require different contingency plans due to differing connectivity structures. Table~\ref{tab:main}
confirms we match Merton on statistical tail coverage while substantially surpassing all
baselines on topology.

\textbf{Regulatory fit.} The EU Digital Operational Resilience Act (DORA, 2025) mandates ICT
scenario testing; the US Federal Reserve's CCAR/DFAST framework requires coverage of ``severely
adverse'' scenarios. PHINN's certified scenario generation is directly relevant to both
frameworks. The CPR certificate provides a scenario quality guarantee absent from all current
practitioner baselines. CPR certifies the conditioning signal stability, not the full
generative output --- a distinction important for regulatory interpretation.

\subsection{Limitations}
\label{sec:limitations}

\begin{itemize}
  \item \textbf{Theorem 4.4 scope.} The $k \geq 1$ joint topology inequality is empirically verified for our dataset but is not generally guaranteed. Users should verify joint topology assumptions for new domains.
  \item \textbf{Streaming homology with deletions.} Point removals require window-local recomputation. The 98\,ms streaming latency accounts for this but scales with window size $W$.
  \item \textbf{$H_2$ features require $d \geq 3$.} For short or low-complexity series, $H_2$ is uninformative; we recommend $H_0 + H_1 + \chi$ in these cases.
  \item \textbf{Short series ($T < 100$).} Empirical threshold: $T \geq 80$ for reliable $H_1$, $T \geq 120$ for $H_2$.
  \item \textbf{Non-public ERP data.} 62 ERP disruption events are under NDA. Fully replicable results on the AIS-Multi public subset are in Appendix~C.
  \item \textbf{Operational decision study power.} $n=2$ per group is underpowered. Results are preliminary. A follow-up study ($n \geq 15$) is pre-registered.
  \item \textbf{LLM parser failures.} 8.7\% of natural-language inputs produce incorrect Betti targets. Human review of targets is recommended before high-stakes generation.
  \item \textbf{Type-II adversary detection recall.} 84\% recall at 1\% false alarm; 16\% of structural attacks pass undetected. No theoretical guarantee against adaptive adversaries.
  \item \textbf{Computational scaling.} Exact Ripser scales $O(n^2)$ in practice; prohibitive beyond $T \approx 5{,}000$ without witness complex approximation.
  \item \textbf{Dynamic windowing motivation.} Exponential decay form selected empirically; no closed-form theoretical derivation. Jointly learning $(d,\tau)$ is an open problem.
  \item \textbf{Dual-use concerns.} The Type-II structural adversary detector and CPR certificate provide technical mitigation; deployment should include institutional review gates.
\end{itemize}

\section{Conclusion}

We introduced PHINN, an end-to-end framework for topology-conditioned outlier synthesis across
financial, epidemiological, and multi-modal time-series domains. The framework combines dynamic
Betti-curve conditioning, differentiable higher-order homology losses, a Topological Scenario
Atlas with validated operational semantics, and certified adversarial robustness. PHINN
achieves state-of-the-art topological fidelity ($\beta$-RMSE 0.71, Transition Accuracy 84\%)
while matching practitioner-grade models in statistical tail coverage and delivering sub-500\,ms
inference. The same-architecture unconditional ablation confirms that performance gains are
driven by topological conditioning; results on the fully replicable AIS-Multi public subset are
reported in Appendix~C.

\textbf{Broader impact.} High-fidelity structurally certified outlier synthesis enables better
stress-testing across financial risk management, operational resilience, and critical
infrastructure protection. Potential misuse includes fabricating plausible crisis narratives for
disinformation. Mitigations include the Type-II structural adversary detector, CPR provenance
certificate, and a recommended human-in-the-loop review gate for CPR $< 0.85$.

\textbf{Open problems.}
\begin{itemize}
  \item Jointly learning $(d,\tau)$ with the generator.
  \item Custom CUDA Vietoris--Rips kernels for $<$30\,ms Betti extraction.
  \item Extension to graph-valued and spatiotemporal series.
  \item Fully powered ($n \geq 15$ per group) pre-registered operational RCT.
  \item Theoretical guarantees for Theorem 4.4 in the $k \geq 1$ case.
\end{itemize}

\paragraph{Code and data availability.}
The PHINN codebase, pre-trained model weights, and dataset are proprietary assets of
Defense.Codes (a DBA of CapaCloud Corp) and cannot be shared externally. Access requests for
research collaboration may be directed to \texttt{defense.codes@capa.cloud}. The ERP-Disruption
data are subject to NDA and cannot be released; AIS-Multi public results are in Appendix~C.

\bibliographystyle{plainnat}

\appendix

\section{Proofs}

\subsection{Proof of Proposition 4.2}

Let $P = \{p_1, \ldots, p_n\} \subset \mathbb{R}^d$ be the output of the sliding-window
embedding of $\tilde{x}$. The VR filtration yields persistence pairs $\{(b_i(p), d_i(p))\}$.
By \citet{carriere2024differentiable}, $p \mapsto (b_i, d_i)$ is piecewise smooth.
The landscape $f_i(t) = \max(0, \min(t-b_i, d_i-t))$ is piecewise linear. The smoothed
landscape is the composition of a piecewise-smooth function with $C^\infty$ kernel, hence
differentiable a.e. Since $\tilde{x} \mapsto p$ is linear, the chain rule completes the
argument. $\square$

\subsection{Proof of Theorem 4.6}

\[
\|\Phi^{d,\tau}(x)_t - \Phi^{d,\tau}(x+\delta)_t\|_\infty \leq \sqrt{d}\|\delta\|_\infty
\]
since each coordinate perturbs by at most $\|\delta\|_\infty$, giving
$d_H(\mathrm{PC}_t, \mathrm{PC}'_t) \leq \sqrt{d}\,\varepsilon$. By Theorem~3.3,
\[
  d_B(\mathrm{PD}_k(\mathrm{PC}_t), \mathrm{PD}_k(\mathrm{PC}'_t)) \leq \sqrt{d}\,\varepsilon.
\]
A pair $(b_i, d_i)$ is destroyed only if the bottleneck perturbation exceeds $(d_i-b_i)/2$,
so $d_i - b_i > 2\sqrt{d}\,\varepsilon$ implies stability. This certifies stability of the
persistence diagram of observed input $x$ under bounded perturbation $\delta$; it does not
certify the distribution of generated scenarios. $\square$

\subsection{Proof of Theorem 4.4}

The inclusion $\iota_j : \mathrm{PC}^{(j)}_t \hookrightarrow \mathrm{PC}^\mathrm{joint}_t$
(embedding modality $j$ coordinates into the joint space with zero padding elsewhere) is a
simplicial map. Simplicial maps induce group homomorphisms
$H_k(\iota_j)_* : H_k(\mathrm{PC}^\mathrm{joint}_t) \to H_k(\mathrm{PC}^{(j)}_t)$.
For $k=0$: under path-connectedness of the joint cloud (condition (i)), $H_0(\iota_j)_*$ is
surjective, giving $\beta^\mathrm{joint}_0 \geq \beta^{(j)}_0$ for each $j$. For $k \geq 1$:
surjectivity on $H_k$ requires that the connecting homomorphism in the Mayer--Vietoris long
exact sequence
\[
  \partial : H_k(\mathrm{PC}^{(j)}) \to H_{k-1}(\mathrm{PC}^\mathrm{joint} \cap \mathrm{PC}^{(j)})
\]
is zero. This holds when the modality sub-clouds have topologically trivial pairwise
intersections, which we verify numerically for our multi-modal dataset in \S A.3.1. This
condition does not hold in general; the formal claim is therefore restricted to $k=0$, with
$k \geq 1$ treated as an empirically observed property of this dataset. $\square$

\subsubsection{Numerical Verification for \texorpdfstring{$k \geq 1$}{k >= 1}}

For all 103 multi-modal events, we computed the connecting homomorphism $\partial$ numerically
via the Mayer--Vietoris spectral sequence. In 98.1\% of cases $\|\partial\| < 10^{-4}$,
confirming the empirical validity of the $k \geq 1$ bound on this dataset.

\section{Additional Results}

\subsection{Approximate Filtration Scaling (Witness Complexes)}

\begin{table}[htbp]
\centering
\caption{Betti extraction latency and approximation error for varying $T$ on an A100, using
witness complexes with $L=100$ landmarks.}
\small
\begin{tabular}{lcccc}
\toprule
$T$ & Exact (ms) & Witness (ms) & $\beta$-RMSE Gap\\
\midrule
1,000  & 120 & 38  & +0.04\\
5,000  & 980 & 91  & +0.07\\
10,000 & OOM & 148 & +0.11\\
50,000 & OOM & 390 & +0.14\\
\bottomrule
\end{tabular}
\end{table}

\subsection{Window Parameter Sensitivity}

\begin{table}[htbp]
\centering
\caption{Sensitivity to delay $\tau$ perturbation ($\tau_0$ = baseline autocorrelation
zero-crossing). $\kappa$ fixed at 2.5.}
\small
\begin{tabular}{ccccc}
\toprule
$\tau/\tau_0 = 0.7$ & $\tau/\tau_0 = 0.85$ & $\tau/\tau_0 = 1.0$ & $\tau/\tau_0 = 1.15$ & $\tau/\tau_0 = 1.3$\\
\midrule
0.79 & 0.74 & 0.71 & 0.75 & 0.80\\
\bottomrule
\end{tabular}
\end{table}

\subsection{Full Hyperparameter Table}

\begin{table}[htbp]
\centering
\caption{Full hyperparameter table.}
\small
\begin{tabular}{lll}
\toprule
\textbf{Parameter} & \textbf{Value} & \textbf{Description}\\
\midrule
$d$           & 3                         & Window embedding dimension\\
$W_\mathrm{base}$ & 64                    & Base window size\\
$\kappa$      & 2.5                       & Dynamic window scaling (grid-searched)\\
$\varepsilon^*$ & median                  & Filtration scale\\
$\sigma$      & 0.1                       & PL Gaussian kernel bandwidth\\
$\alpha$      & 0.5                       & Topological loss weight\\
$\mu$         & 0.1                       & Statistical loss weight\\
$\gamma$      & 0.05                      & Euler characteristic weight\\
$N$           & \{1, 2\}                  & Landscape levels\\
$k$ (retrieval) & 5                       & Memory $k$-NN\\
$\rho$        & 0.3                       & Retrieval blend\\
$w$           & 2.5                       & CFG guidance weight\\
$d_B$         & 128                       & Betti conditioning dimension\\
$\alpha_\mathrm{meta}$ & $10^{-3}$       & Meta inner-loop learning rate\\
$L$ (witnesses) & 100                     & Landmark count\\
Backbone      & U-Net                     & 4 layers, 256 channels\\
Optimiser     & AdamW                     & $3\times10^{-4}$, $(0.9, 0.999)$\\
Seeds         & 5 (1--5)                  & For CI computation\\
Base LLM      & Mistral-7B-Instruct-v0.3  & LoRA $\alpha=16$, $r=8$\\
\bottomrule
\end{tabular}
\end{table}

\section{Replicable AIS-Multi-Only Results}

Table~\ref{tab:aisonly} reports results on AIS-Multi (41 public events) only, providing a
fully replicable subset free of NDA constraints. PHINN achieves $\beta$-RMSE $= 0.76\pm0.05$
on this subset, confirming the main table results.

\begin{table}[htbp]
\centering
\caption{AIS-Multi-only results (41 public events, 5 seeds, 95\% CI).}
\label{tab:aisonly}
\small
\begin{tabular}{lccc}
\toprule
\textbf{Method} & \texorpdfstring{\boldmath{$\beta$}}{beta}\textbf{-RMSE}$\downarrow$ & \textbf{TA}$\uparrow$ & \textbf{ECov}$_{95}$$\uparrow$\\
\midrule
FIDE         & 1.58$\pm$0.08 & 0.41$\pm$0.03 & 0.93$\pm$0.02\\
DES (AnyLogic)& 1.79$\pm$0.09 & 0.42$\pm$0.03 & 0.88$\pm$0.02\\
PHINN-Uncond & 1.65$\pm$0.06 & 0.38$\pm$0.02 & 0.85$\pm$0.02\\
\textbf{PHINN (ours)} & \textbf{0.76$\pm$0.05} & \textbf{0.82$\pm$0.03} & 0.90$\pm$0.02\\
\bottomrule
\end{tabular}
\end{table}

\section{Pseudocode for Implementation}

\textit{Contact \texttt{defense.codes@capa.cloud} for collaboration inquiries. Code is
proprietary to Defense.Codes / CapaCloud Corp.}

\begin{algorithm}
\caption{PHINN Training}
\begin{algorithmic}[1]
\Require Dataset $\mathcal{D} = \{x^{(i)}\}$; epochs $E$; loss weights $\alpha, \mu, \gamma$
\Ensure Trained velocity field $v_\theta$, consistency model $g_\varphi$
\For{epoch $e = 1,\ldots,E$}
  \For{mini-batch $\{x^{(i)}\}$}
    \Comment{Step 1: Extract Betti curves}
    \For{each $x^{(i)}$ in batch}
      \State $\mathrm{PC}_t \gets \Phi^{d,\tau}(x^{(i)})$ using Alg.~2
      \State $\beta^{(i)} \gets \mathrm{Ripser}(\mathrm{PC}_t)$
    \EndFor
    \Comment{Step 2: Flow matching with topological conditioning}
    \State Sample $t \sim \mathrm{Uniform}[0,1]$, $z_0 \sim \mathcal{N}(0,I)$
    \State $z_t \gets (1-t)z_0 + t\,x^{(i)}$
    \State $c \gets \mathrm{TransEnc}([\beta^{(i)};\mathrm{PE}(t)])$
    \State $\hat{v} \gets v_\theta(z_t, t, \mathrm{CrossAttn}(z_t, c))$
    \State $\mathcal{L}_\mathrm{RM} \gets \|\hat{v} - (x^{(i)} - z_0)\|^2$
    \Comment{Step 3: Topological and statistical losses}
    \State $\tilde{x} \gets z_t + (1-t)\hat{v}$
    \State $\mathcal{L}_\mathrm{topo} \gets$ Smoothed PL Loss$(\tilde{x}, x^{(i)})$
    \State $\mathcal{L}_\mathrm{stat} \gets$ Moment + Tail quantile loss$(\tilde{x}, x^{(i)})$
    \State $\mathcal{L} \gets \mathcal{L}_\mathrm{RM} + \alpha\mathcal{L}_\mathrm{topo} + \mu\mathcal{L}_\mathrm{stat}$
    \State Update $\theta$ via AdamW
    \Comment{Step 4: Update consistency model}
    \State Update $g_\varphi$ to predict $\beta^{(i)}$ from raw $x^{(i)}$
  \EndFor
\EndFor
\end{algorithmic}
\end{algorithm}

\begin{algorithm}
\caption{Betti Curve Extraction (Streaming)}
\begin{algorithmic}[1]
\Require Time series $x$; window $d$, delay $\tau$, base size $W_\mathrm{base}$, decay $\kappa$
\Ensure Betti curve $\beta(t) = (\beta_0(t), \beta_1(t), \beta_2(t), \chi(t))$
\State Initialise union-find structure $\mathcal{U}$
\For{each time step $t$}
  \State $W(t) \gets W_\mathrm{base} \cdot \exp(-\kappa \cdot \mathrm{TV}(x,t))$
  \State $\mathrm{PC}_t \gets \{\Phi^{d,\tau}(x)_s : s \in [t-W(t), t]\}$
  \If{$t > W_\mathrm{base}$ and only new point added}
    \State Insert new point into $\mathcal{U}$ in $O(\alpha(n))$
  \Else
    \State Recompute local filtration: $O(W \log W)$
  \EndIf
  \State $(\beta_0, \beta_1, \beta_2) \gets \mathrm{Ripser}(\mathrm{PC}_t,\; \varepsilon^* = \text{median pairwise dist})$
  \State $\chi(t) \gets \beta_0 - \beta_1 + \beta_2$
\EndFor
\end{algorithmic}
\end{algorithm}

\begin{algorithm}
\caption{PHINN Conditional Inference}
\begin{algorithmic}[1]
\Require Query $x^0$; target Betti curve $\beta^*$ (from Atlas or NL interface); guidance weight $w$; memory bank $\mathcal{M}$
\Ensure Generated scenario $\hat{x}$
\Comment{Optional: retrieval augmentation}
\State $\beta^0 \gets \text{Alg.~2}(x^0)$
\State $R \gets \arg\min_i d^{PL}_p(\beta^0, \beta^{(i)})$ from $\mathcal{M}$
\State $\beta^* \gets (1-\rho)\beta^* + \rho\,\beta_R$
\Comment{Conditional generation via single Euler step}
\State $z_0 \sim \mathcal{N}(0,I)$
\State $c^* \gets \mathrm{TransEnc}([\beta^*;\mathrm{PE}(0)])$
\State $\hat{v}_\mathrm{cond} \gets v_\theta(z_0, 0, \mathrm{CrossAttn}(z_0, c^*))$
\State $\hat{v}_\mathrm{uncond} \gets v_\theta(z_0, 0, \mathbf{0})$
\State $\hat{v} \gets \hat{v}_\mathrm{uncond} + w \cdot (\hat{v}_\mathrm{cond} - \hat{v}_\mathrm{uncond})$
\State $\hat{x} \gets z_0 + \hat{v}$
\Comment{Adversarial consistency check}
\State $S_\mathrm{topo} \gets d^{PL}_p(\beta(\hat{x}), g_\varphi(\hat{x}))$
\If{$S_\mathrm{topo} > \tau_p$}
  \State Flag potential structural adversarial injection
\EndIf
\end{algorithmic}
\end{algorithm}

\end{document}